\documentclass[10pt,twocolumn,letterpaper]{article}

\usepackage{cvpr}              %
\usepackage{graphicx}
\usepackage{amsmath}
\usepackage{amssymb}
\usepackage{booktabs}
\usepackage{comment}
\usepackage{mathtools}
\usepackage{scalerel}
\usepackage[dvipsnames]{xcolor}

\usepackage[pagebackref,breaklinks,colorlinks=true,linkcolor=black,anchorcolor=black,citecolor=black,filecolor=black,menucolor=black,runcolor=black,urlcolor=black]{hyperref}
\AtBeginDocument{\hypersetup{pdfborder={0 0 0}}}

\usepackage[capitalize]{cleveref}
\crefname{section}{Sec.}{Secs.}
\Crefname{section}{Section}{Sections}
\Crefname{table}{Table}{Tables}
\crefname{table}{Tab.}{Tabs.}

\def\cvprPaperID{*****} %
\def\confName{CVPR}
\def\confYear{2022}

\newcommand{\toolname}{\textsc{VisCUIT}\xspace}
\newcommand{\subgroupwindow}{\textit{Subgroup Panel}\xspace}
\newcommand{\neuronpathwindow}{\textit{Neuron Activation Panel}\xspace} 
\newcommand{\gradcamwindow}{Grad-CAM Window\xspace}
\newcommand{\gradcamwindows}{Grad-CAM Windows\xspace}
\newcommand{\conceptwindow}{Neuron Concept Window\xspace} 
\newcommand{\conceptwindows}{Neuron Concept Windows\xspace}

\begin{document}

\title{\toolname: Visual Auditor for Bias in CNN Image Classifier}

\author{
Seongmin Lee \\
Georgia Institute of Technology\\
Atlanta, Georgia\\
{\tt\small seongmin@gatech.edu}
\and 
Zijie J. Wang \\
Georgia Institute of Technology\\
Atlanta, Georgia\\
{\tt\small jayw@gatech.edu}
\and 
Judy Hoffman\\
Georgia Institute of Technology\\
Atlanta, Georgia\\
{\tt\small judy@gatech.edu}
\and 
Duen Horng (Polo) Chau \\
Georgia Institute of Technology\\
Atlanta, Georgia\\
{\tt\small polo@gatech.edu}
}
\maketitle

\begin{figure*}[h]
    \centering
    \includegraphics[width=2 \columnwidth]{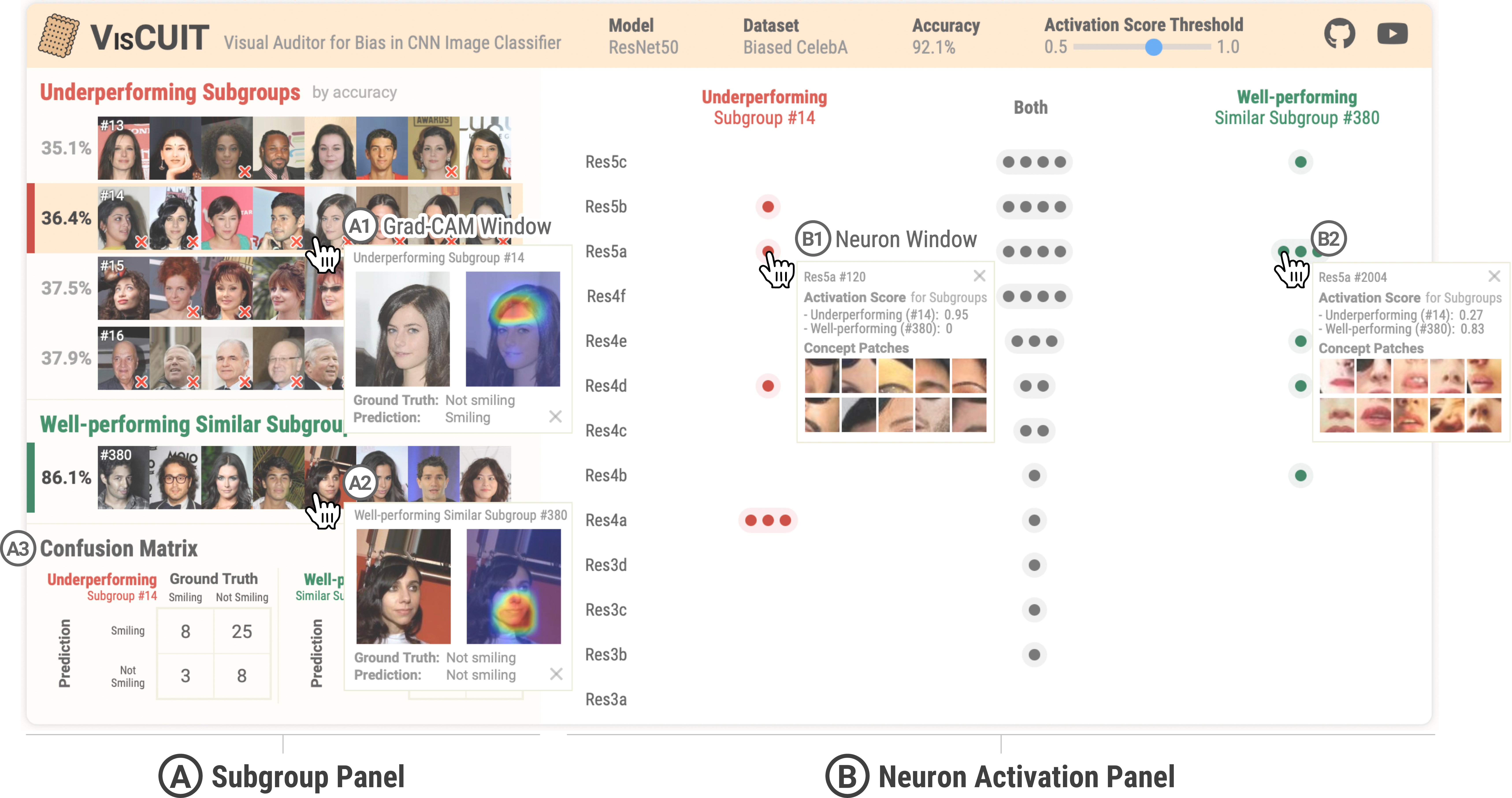}
    \caption{
    \toolname 
    reveals \textit{how} and \textit{why} a CNN image classifier is biased. 
    Our user Jane
    trains a classifier
    using the biased CelebA dataset,
    which has high co-occurrence of the attribute \textit{black hair} and the label \textit{smiling},
    to observe how the training data affects model predictions.
    She hypothesizes that the model would use the attribute \textit{black hair} to predict \textit{smiling}
    and launches \toolname to verify her hypothesis.
    (A) \subgroupwindow displays underperforming data subgroups
    found by UDIS~\cite{KrishnakumarPSH21}.
    Jane figures out that several underperforming subgroups
    consist of people with \textit{black hair}. 
    To see whether the model indeed uses the attribute \textit{black hair} for predictions,
    Jane clicks on subgroup \#14, and  
    \toolname displays subgroup \#380,
    which is similar to \#14 in terms of the last-layer feature vectors from the model
    but has high accuracy.
    Clicking on an image in each of those subgroups
    brings up a \gradcamwindow, 
    which shows that the classifier attends to
    (A1) \textit{forehead} (near \textit{hair}, irrelevant to \textit{smiling}) for the subgroup \#14
    and
    (A2) \textit{mouth} (relevant to \textit{smiling}) for the subgroup \#380.
    (A3) Confusion matrices quantitatively summarize such misclassifications 
    that many \textit{not smiling} \textit{black-haired} people
    are wrongly classified as \textit{smiling}.
    Jane is now certain %
    that the classifier uses the attribute \textit{black hair} for predicting \textit{smiling} 
    and therefore often misclassifies \textit{black-haired} people.
    (B) The \neuronpathwindow enables users to understand 
    which neurons and concepts are responsible for misclassifications,
    by organizing the neurons in the model into 3 columns:
    the left column for the neurons highly activated only by underperforming subgroup,
    the right only by well-performing subgroup,
    and the middle by both.
    Clicking on a neuron displays a \conceptwindow,
    which reveals that
    (B1, B2)
    the subgroups \#14 and \#380
    activate the neurons for the area near
    \textit{forehead} and \textit{mouth}, respectively.
    }
    \label{fig:viscuit_ui}
\end{figure*}

\begin{abstract}

CNN image classifiers are widely used, thanks to their efficiency and accuracy.
However, they can suffer from biases that impede their practical applications.
Most existing bias investigation techniques 
are either inapplicable to general image classification tasks
or require significant user efforts in perusing all data subgroups to
manually specify which data attributes to inspect.
We present \toolname,
an interactive visualization system
that reveals
\textit{how} and \textit{why} a CNN classifier is biased.
\toolname visually summarizes the subgroups on which the classifier underperforms
and helps users discover and characterize the cause of the underperformances
by revealing image concepts responsible for activating neurons that contribute to misclassifications.
\toolname runs in modern browsers and is open-source,
allowing people to easily access and extend the tool to other model architectures and datasets.
\toolname is available at the following public demo link: \url{https://poloclub.github.io/VisCUIT}.
A video demo is available at
\url{https://youtu.be/eNDbSyM4R_4}.

\end{abstract}

\section{Introduction}
\label{sec:100introduction}
Recently,
data classification algorithms are widely used for practical applications,
such as
face recognition~\cite{ParmarM14,TaigmanYRW14,WangD21a},
autonomous driving~\cite{PrabhakarKNK17,FujiyoshiHY19},
and clinical trials~\cite{Topol19,ShahKKGHLRS19,ZoabiDS21}.
Despite the fact that visual models outperform humans in some circumstances~\cite{Buetti-DinhGBIHCBPWSVD19},
several works have found that these classifiers are often biased with disparate performance across data subgroups
~\cite{AngwinLMK16,BuolamwiniG18,KrishnapriyaAVKB20,MehrabiMSLG21}.
Exploiting the biased classifiers for critical purposes can cause
unintentional fairness violation and huge societal problems~\cite{GarvieBF16,OsobaW17,ZweigWK18,Washington19}.

Likewise,
image classifiers based on deep convolutional neural networks (CNN),
which have achieved state-of-the-art performance in various areas~\cite{KrizhevskySH12,SzegedyLJSRAEVR15,HeZRS16,SharmaJM18,SultanaSD19},
often suffer from biases~\cite{KrishnapriyaAVKB20}.
To facilitate real-world applications of the state-of-the-art techniques,
there have been attempts to
understand~\cite{AlbieroSVZKB20,DrozdowskiRDDB21,CavazosPCO21}
and
mitigate~\cite{WangZYCO19,WangD20,DharGSCC20,Gong0021} 
the biases in CNN classifiers.
However,
most existing methods require humans to specify the attributes on which to audit the classifiers.
As people tend to focus more on sensitive attributes (e.g., race, gender),
less sensitive attributes (e.g., wearing glasses, hair color) that can correlate with biases and degrade the overall performance
are easily missed.
Existing approaches 
assume availability of additional attributes other than the class label for each image;
thus, datasets without any additional attributes %
cannot be analyzed using these methods.

Krishnakumar et al.~\cite{KrishnakumarPSH21}  proposed UDIS,
which automatically detects the data subgroups,
on which a CNN classifier underperforms.
While UDIS does not require additional attribute labels, the approach produces a large number of potentially biased subgroups which may or may not align with semantic concepts. This leads to ambiguous findings even after substantial manual inspection.
Moreover, 
most aforementioned bias investigation approaches
detect the source of biases in classifiers,
primarily focusing on
their training datasets, 
not how the neurons in the classifier are activated and generate biased outputs~\cite{BuolamwiniG18,AlbieroSVZKB20,Derman21}.

In this paper,
we present \toolname,
an interactive visualization system
that reveals \textit{how} and \textit{why} a CNN image classifier is biased,
 without requiring users to pre-determine which attributes to inspect.
\toolname's major contributions include:
\begin{itemize}
    \item 
    \textbf{Visual summarization of the undeperforming subgroups.}
    \toolname highlights the data subgroups generated by  UDIS~\cite{KrishnakumarPSH21} on which a CNN classifier underperforms.
    This allows users to understand \textit{how} the classifier is biased,
    not limiting the bias factors to the sensitive attributes.
    As \toolname summarizes the undeperforming subgroups as a list,
    users can easily characterize each subgroup.
    For each undeperforming subgroup,
    \toolname also displays its most similar subgroup with high accuracy, 
    based on Euclidean distance in the feature space, 
    enabling users to gain insights into the deviant features responsible for the biases ~\cite{KrishnakumarPSH21}.

    \item 
    \textbf{Visual bias attribution  in CNN image classifiers.}
    \toolname demonstrates \textit{why} %
    a CNN classifier underperforms on each subgroup,
    by revealing image concepts responsible for activating neurons
    that contribute to the underperformances.
    Users can observe how the classifier is activated differently by the underperforming and well-performing subgroups,
    focusing on the high-level concepts in images.
    Moreover, for each image, 
    \toolname displays \gradcamwindow,
    which visually highlights features in an input image deemed relevant for classification~\cite{SelvarajuCDVPB17}.%

    \item 
    \textbf{Open-sourced, web-based implementation.}
    \toolname runs directly in modern browsers and is open-source,\footnote{Code: \url{https://github.com/poloclub/VisCUIT}}
    allowing people to easily access and extend the tool to other model architectures and datasets.
    Figure~\ref{fig:viscuit_ui} illustrates the user interface of \toolname.
    \toolname is available at the following public demo link:
    \url{https://poloclub.github.io/VisCUIT}.
    A video demo is available at \url{https://youtu.be/eNDbSyM4R_4}.
\end{itemize}

\section{Related Works}
\label{sec:200related}
\subsection{Identification of Biases in Algorithms}

There have been many efforts to identify
biases (i.e. the disparity in the performance among the data subgroups)
in the state-of-the-art algorithms.
Many of the face recognition algorithms 
have been proven to include racial and gender biases~\cite{ZouS18,BuolamwiniG18,AlbieroSVZKB20,CavazosPCO21}.
Lambrecht et al.~\cite{LambrechtT19} and
Angwin et al.~\cite{AngwinLMK16}
revealed that
the advertisement recommendation systems
and legal decision making software
are also biased against specific ethnicity or gender.
However, all these approaches require people to predefine protected attributes to audit the algorithms;
therefore, only few sensitive factors (e.g., race, gender) are considered. %
Moreover, these methods are inapplicable to the image classification tasks
whose datasets do not contain any additional attributes,
such as ethnicity and gender,
other than the class labels.

To analyze biases in general image classifiers,
Singh et al.~\cite{SinghMGLFG20} investigated the co-occurence between objects and their contexts for each category
and attempted to decorrelate them to reduce classifiers' dependency on the contexts.
UDIS~\cite{KrishnakumarPSH21},
which is developed to generate the image subgroups without human guides or additional attributes,
clustered images based on the last-layer feature vectors from classifiers
and extracted the subgroups with low accuracies.
However,
it is hard to define how a classifier is biased using these methods
since numerous subgroups are generated 
and the characteristics of each subgroup is often unclear.
Furthermore,
most existing approaches~\cite{BuolamwiniG18,AlbieroSVZKB20,Derman21} %
argue that 
the skewness in training datasets %
is the major source of biases
but do not inspect how the neurons in the classifiers are activated
and generate biased outputs.
Different from the existing methods,
\toolname visually summarizes the discovered subgroups to allows users to 
easily define each underperforming subgroup.
Also,
\toolname reveals which neurons and image concepts are responsible for making predictions for each subgroup,
so that users can learn more about why the classifier underperforms on some subgroups.

\subsection{Bias Analysis Toolkits}

Since bias can hugely affect various people's lives,
the toolkits
to help people without much background knowledge understand algorithmic biases
have been actively developed.
FairML~\cite{Adebayo16} quantifies the relative significance of the inputs to a predictive model
to evaluate the model's fairness.
While Aequitas~\cite{SaleiroKSAHLG18} 
enables users to easily measure fairness of algorithms
using various metrics,
AI Fairness 360~\cite{BellamyDHHHKLMM19}
integrates a number of state-of-the-art techniques for algorithmic biases,
including bias assessment metrics, bias mitigation algorithms, and bias explanations.
FairVis~\cite{CabreraEHKMC19}
allows users to generate and explore data subgroups
based on their domain knowledge
and suggests relevant subgroups.
However,
these approaches are applicable only for the datasets with abundant well-defined attributes (e.g. tabular data),
and therefore cannot handle the models associated with image data
unless additional attributes for the data are provided.

\subsection{CNN Analysis Techniques}

A growing body of research has proposed techniques
to help people interpret the behaviors of CNN models.
Earlier CNN interpretation approaches %
have made input-level explanations,
which aim to reveal 
the features in inputs with major contribution to model behaviors~\cite{SimonyanVZ13,SelvarajuCDVPB17}.
However, 
these approaches do not demonstrate which neurons in CNN models are responsible for the model behaviors.
Recently, several methods propose neuron-level explanations~\cite{fong2018net2vec,HohmanPRC20,ParkDDWSHC21}.
In parallel,
some research attempts to interpret adversarial attack in CNNs~\cite{das2020bluff,cantareira2021explainable,ma2021understanding},
hyperparameter tuning~\cite{optuna_2019,joo2021guided,jin2021keras},
and model selection~\cite{mostafa2021visualizing}.
\toolname focuses on \textit{biases} in CNN models
and investigates neuron activations in the model
to understand why the model generates biased outputs.

\section{System Design and Implementation}
\label{sec:300design}

\subsection{Overview}
\label{subsec310}

\noindent
\textbf{User Interface.}
\toolname aims to reveal \textit{how} and \textit{why} a CNN image classifier is biased.
\toolname consists of 
the \subgroupwindow (Figure~\ref{fig:viscuit_ui}A) and the \neuronpathwindow (Figure~\ref{fig:viscuit_ui}B). 
The \subgroupwindow
displays image subgroups on which the CNN classifier underperforms
and allows users to select a subgroup to explore.
For a selected underperforming subgroup,
the \subgroupwindow shows
a well-performing subgroup similar to the selected underperforming subgroup, where similarity is determined based on the last-layer feature vectors from the classifier.
At the bottom of the \subgroupwindow, the confusion matrices of the two subgroups are displayed.
The \neuronpathwindow (Figure~\ref{fig:viscuit_ui}B)
helps users discover and characterize the
cause of the underperformances, by revealing image concepts responsible for activating neurons that contribute to the subgroups' predictions.

\smallskip
\noindent
\textbf{Dataset and Open-source System Implementation.}
In our demo,
we investigate the ResNet50~\cite{HeZRS16} classifier
that has been trained with the biased CelebA~\cite{LiuLWT15,KrishnakumarPSH21} dataset
to predict whether a person in an image is \textit{smiling}
and has achieved an accuracy of 92.1\%.
To verify the validity of \toolname,
we intentionally increase the co-occurrence of the attribute \textit{black hair} and the label \textit{smiling}, 
so that the classifier would more likely use image features related
to \textit{black hair} to predict \textit{smiling};
and \toolname would identify such biases.
\toolname is open-source,
and can be easily extended to support other model architectures and datasets.
We have implemented \toolname using the standard HTML/CSS/JavaScript web technology stack
and the D3.js~\cite{BostockOH11} visualization library.
CNN model training and inference
are all implemented using PyTorch~\cite{paszke2019pytorch}.

\subsection{Subgroup Panel}
\label{subsec320}
\noindent
\textbf{Underperforming Subgroups.}
The \subgroupwindow (Figure~\ref{fig:viscuit_ui}A),
shows a list of underperforming subgroups,
whose accuracies are much lower than the model's overall accuracy of 92.1\%.
We adopt the UDIS~\cite{KrishnakumarPSH21}  subgroup discovery algorithm to identify 
these underperforming subgroups,
which works by clustering the images
based on their feature vectors from the last layer of the classifier,
and then collecting the clusters with accuracies
lower than half of the overall accuracy.
Each subgroup's accuracy and images are displayed,
and the subgroups are sorted by accuracy.
An image incorrectly predicted by the classifier
is marked with a small red cross (\scalerel*{\includegraphics{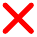}}{Q}).

\smallskip
\noindent
\textbf{Most Similar Subgroup with High Accuracy.}
When the user clicks an underperforming subgroup, 
\toolname displays its most similar subgroup with high accuracy, 
based on Euclidean distance in the feature space, 
enabling users to gain insights into the deviant features responsible for the biases ~\cite{KrishnakumarPSH21}.
We call this subgroup the \textit{``well-performing similar subgroup.''}
In more detail, 
to assess the similarity between subgroups,
we compute the vector embedding of each subgroup,
by averaging the last-layer feature vectors from the classifier of all the images in the subgroup.
Using the obtained subgroup embeddings,
we evaluate the Euclidean distances between subgroups
and regard the well-performing subgroup closest to the selected underperforming subgroup
as its most similar well-performing subgroup.
The well-performing subgroup is summarized in the same format as the underperforming subgroup,
displaying its accuracy and images.

\begin{figure}
    \centering
    \includegraphics[width=\columnwidth]{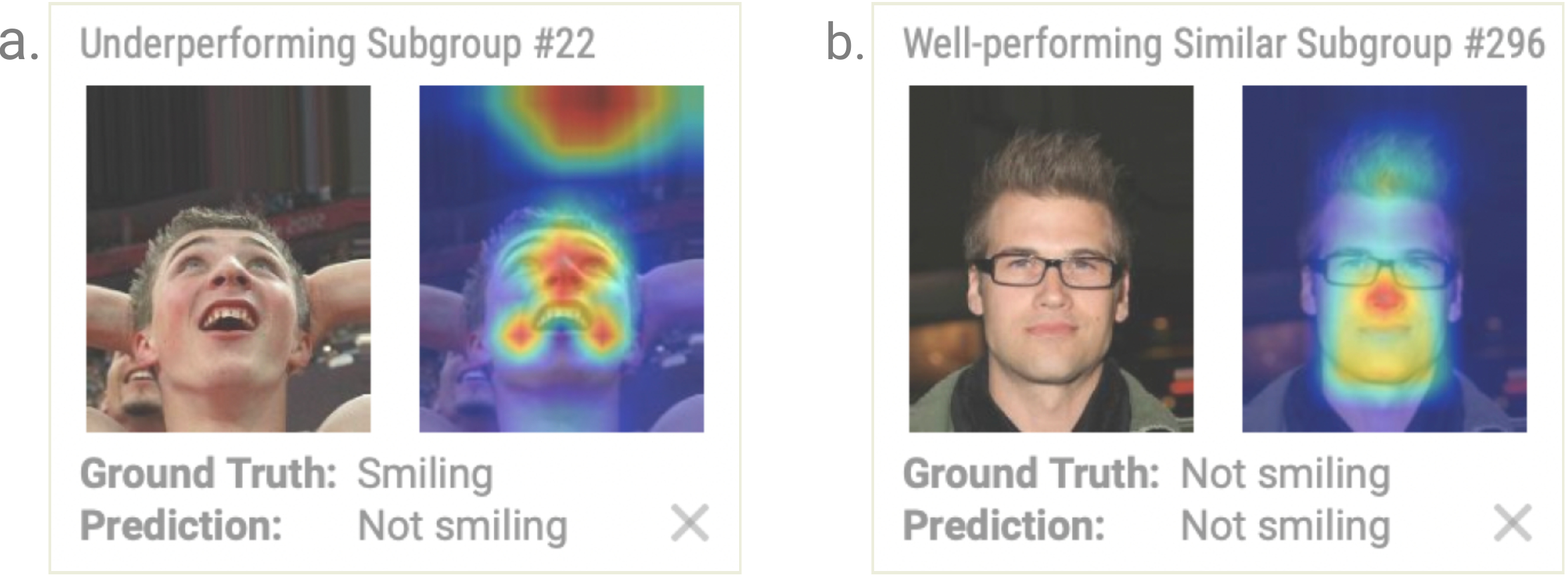}
    \caption{
    The \textbf{\gradcamwindow} helps users understand the reasons for misclassifications; 
    it is displayed when a user clicks on an image.
    (a) \gradcamwindow for an image 
    in the underperforming subgroup \#22
    reveals that the model attends to background areas not relevant to facial expressions. 
    (b) \gradcamwindow for an image 
    in the well-performing subgroup \#296
    reveals that the model attends to the face as expected.
    }
    \label{fig:gradcam}
\end{figure}

\smallskip
\noindent
\textbf{\gradcamwindow{}.}
When user clicks an image
in the selected underperforming subgroup
or its well-performing similar subgroup,
a \gradcamwindow
pops up (Figure~\ref{fig:gradcam}).
The window contains the selected image's prediction results and
Grad-CAM~\cite{SelvarajuCDVPB17} saliency visualizations.
Grad-CAM is one of the most popular methods that visually highlights features in an input image deemed relevant for classification;
using Grad-CAM,
users can more easily 
understand why an image is incorrectly classified~\cite{SelvarajuCDVPB17}.

\smallskip
\noindent
\textbf{Subgroup Confusion Matrix.}
The bottom of the \subgroupwindow
shows the confusion matrices of the selected underperforming subgroup and its similar well-performing subgroup
to summarize
the prediction results within those subgroups.
It helps users 
more easily assess the types of classification errors and their distributions across the class labels. 

\subsection{Neuron Activation Panel}
\label{subsec330}
The \neuronpathwindow (Figure~\ref{fig:viscuit_ui}B)
helps users discover the cause of the underperformances,
by revealing image concepts responsible for activating neurons that contribute to misclassifications
of the selected underperforming subgroup.

\smallskip
\noindent
\textbf{Neuron Activations.}
The \neuronpathwindow
displays the highly activated neurons
for the selected underperforming subgroup and its similar well-performing subgroup.
To reveal how the two subgroups diverge in the classifier,
we organize the neurons into 3 columns:
\textit{Underperforming Subgroup}, \textit{Both}, and \textit{Well-performing Similar Subgroup}.
The neurons in the columns
\textit{Underperforming Subgroup} and \textit{Well-performing Similar Subgroup}
are activated only by the underperforming and well-performing similar subgroup, respectively,
while the neurons in the column \textit{Both} are activated by both of the subgroups.
To help users more easily assess each layer's contribution to the predictions, the neurons are organized vertically based on their layers in the classifier.

For each neuron,
we evaluate neuron activation score,
which is the baseline for differentiating \textit{highly activated neurons} from others.
The neuron activation score is evaluated
based on the neuron importance measurement method in Summit~\cite{HohmanPRC20}.
For each image,
each neuron's maximum activation across spatial locations
is considered as the activation value of the neuron for the image.
Then, for each layer,
we extract the neurons with the highest activation values
until the sum of the extracted neurons' activation values exceeds 3\% of the layer's total activation value,
and we consider the extracted neurons as the \textit{highly activated neurons} for the image.
We identify \textit{highly activated neurons} for all the images in a subgroup,
and for each of the neurons,
we calculate the proportion of the images in the subgroup
that have the neuron as their \textit{highly activated neuron}.
The proportion is used as the neuron's activation score for the subgroup.

The header above the \neuronpathwindow
displays a slider
to adjust the threshold for neuron activation score.
When user increases the threshold,
the neurons,
whose activation scores for either
the selected underperforming subgroup or its similar well-performing subgroup
are lower than the threshold,
are relocated or filtered out.
The threshold ranges from 0.5 to 1.0;
we set the lower bound to 0.5
to prevent numerous neurons from appearing and overwhelming users.

\begin{figure}
    \centering
    \includegraphics[width=0.95 \columnwidth]{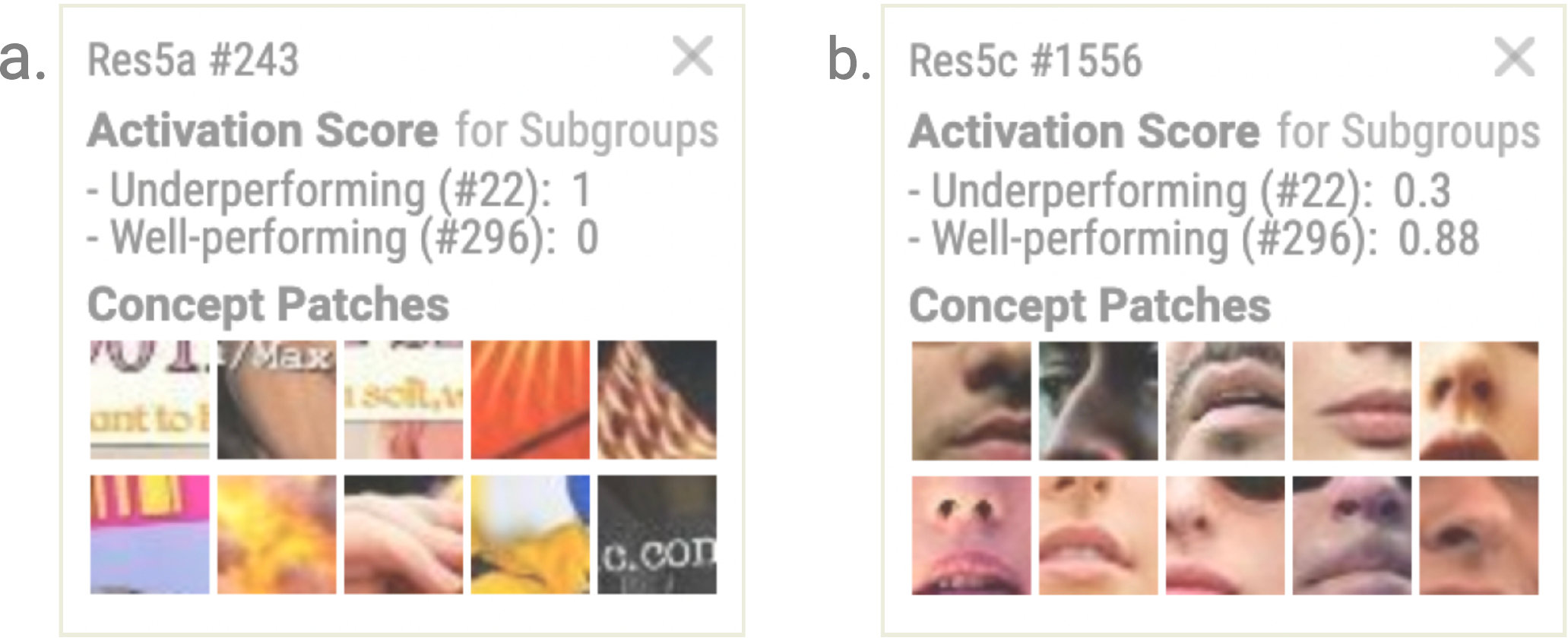}
    \caption{
    The \textbf{\conceptwindow{}} helps users identify image concepts responsible for activating neurons 
    that significantly contribute to model prediction. %
    (a) \conceptwindow{} for a neuron highly activated only by the underperforming subgroup \#22
    shows that \textit{text} and \textit{non-facial textures}, both irrelevant to predicting \textit{smiling}, 
    have the major contribution to the misclassification, 
    with the high activation score of 1.
    (b) \conceptwindow{} for a neuron 
    highly activated by the well-performing subgroup \#296
    shows the expected image features, such as \textit{mouths} and \textit{noses}.
    The \conceptwindow displays when the user clicks on a neuron in the \neuronpathwindow.
    }
    \label{fig:neuronconcept}
\end{figure}

\smallskip
\noindent
\textbf{\conceptwindow.}
When a neuron is clicked, \toolname shows a \conceptwindow,
which contains
the neuron's activation scores for the underperforming and well-performing subgroups
and concept patches (Figure~\ref{fig:neuronconcept}).
This window helps users
understand how much and why each of the neurons have been activated by each subgroup.
We generate the concept patches
based on the existing methods~\cite{HohmanPRC20,ParkDDWSHC21};
for each neuron,
we get the 10 images that activate the neuron the most over the entire dataset.
Then, for each of the images, %
we randomly generate 32 masks,
each of which is for a square concept patch (30px by 30px).
We separate the square areas of different masks 
to be at least 5px apart from each other
to promote diversity among the concept patches.%
We then 
input all the concept patches to the classifier
and observe how the neurons in the classifier are activated.
For each neuron,
10 concept patches that activate the neuron the most are considered as the neuron's concept patches.

\smallskip
\noindent
\textbf{Neuron Clustering.}
It is known that
some neurons in CNN have redundancy
and are activated by similar concepts~\cite{JaderbergVZ14,wen2016learning,he2017channel,duggal2021cup}.
To help users identify such redundancy
and focus on distinct concepts,
when a neuron is hovered,
we highlight the neurons that 
have similar sets of concept patches
with the hovered neuron.
Inspired by the neuron clustering method in \cite{ParkDDWSHC21},
we identify the neuron clusters activated by the same concepts.
We train an additional model, which is based on the ResNet50 architecture,
that takes concept patches as inputs 
and outputs a vector for each concept patch
to maximize the inner products between the vectors of the concept patches for the same neurons.
We randomly sample 10,000 concept patch pairs,
each of which consists of two patches from the same neurons,
to generate training dataset;
for negative sampling,
we additionally sample 10,000 concept patch pairs,
each of which is two patches from two different neurons.
The objective function to be minimized is
\begin{equation}
    -\sum_{\mathclap{\substack{V_i,V_j\in\\ \text{Same Neuron}}}}{\;\log(V_i\cdot V_j)}
    -\sum_{\mathclap{\substack{V_i',V_j'\in\\ \text{Different Neurons}}}}{\;\log(1-V_i'\cdot V_j')}
\end{equation}
where $V_i, V_j$ and $V_i', V_j'$ are the normalized vectors of the concept patches from the same and different neurons, respectively.
We initialize the model with the classifier that we are investigating
and train for 10 epochs
using the SGD optimizer with learning rate of 0.0001.
After training the model,
we iterate each of the \textit{highly activated neurons} in the classifier
to compute the inner products between
the vectors for the neuron's concept patches
and the vectors for the concept patches sampled from each neuron cluster.
Among the neuron clusters,
a neuron is added to the cluster that yields the maximum inner product
if the inner product value is greater than the preset threshold 0.9;
otherwise, we generate a new cluster with the neuron as the only element.
We set the threshold for adding a neuron to a cluster to 0.9
to minimize the error
that any two neurons for different concepts are grouped to the same cluster.

\section{Usage Scenario}
\label{sec:400design}

\subsection{Bias Characterization}

\begin{figure}
    \centering
    \includegraphics[width=0.95\columnwidth]{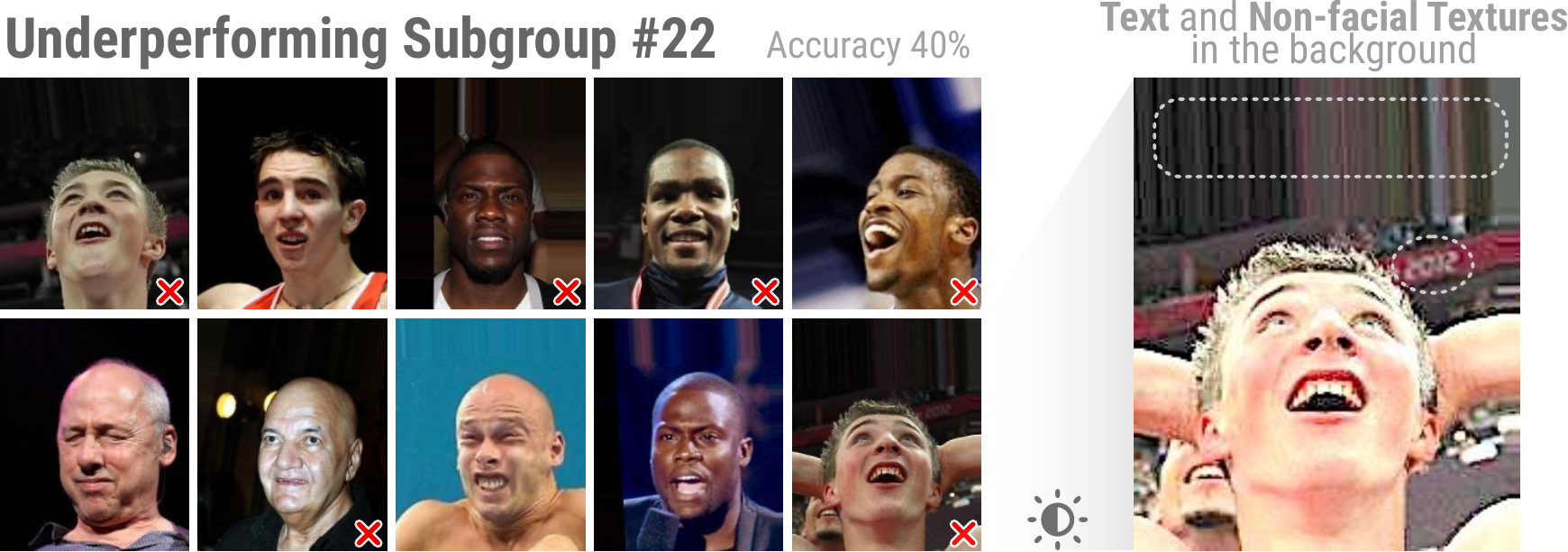}
    \caption{
    Images in the underperforming subgroup \#22.
    It is hard to define the characteristics of the subgroup \#22 at the first glance.
    Using \toolname,
    our user Henry gets hints that there would be
    text and non-facial textures in the background 
    and verifies it by brightening the image.
    Henry concludes that the classifier underperforms on the \textit{images of smiling athletes in stadium}.
    }
    \label{fig:subgroup22}
\end{figure}

\toolname helps users characterize the images where the classifier underperforms.
For example,
a hypothetical machine learning engineer Henry is using \toolname
to investigate the ResNet50 classifier that predicts whether a person in an image is \textit{smiling}.
While scrolling down the list of underperforming subgroups,
Henry is curious about
subgroup \#22 with an accuracy of 40\% (Figure~\ref{fig:subgroup22})
since
he finds it hard to define the common characteristics of the images in the subgroup.
Henry decides to look into the subgroup \#22,
wishing to %
figure out what kinds of images consist of the subgroup \#22
to clarify how the classifier is biased.

As Henry clicks the subgroup \#22,
\toolname displays
the well-performing subgroup \#296,
which is similar to the subgroup \#22 in terms of the feature vectors from the classifier,
confusion matrices of the subgroups \#22 and \#296,
and the neurons in the classifier activated by the two subgroups.
Henry first clicks on the images in the subgroups \#22 and \#296
to examine the \gradcamwindows and compare the two subgroups
(Figure~\ref{fig:gradcam}).
From the \gradcamwindows,
Henry figures out that the classifier anomalously attends to the background areas,
which look not pertinent to \textit{smiling} at all,
for the images in the underperforming subgroup \#22.
Also,
from the confusion matrices,
he learns that the classifier predicts all the images in the subgroup \#22 as \textit{not smiling},
even though more than half of them are actually \textit{smiling}.
Wondering why the background areas are attended by the classifier,
he moves on to the \neuronpathwindow to scrutinize the neuron activations.

Since there are many neurons in the \neuronpathwindow,
Henry increases the activation score threshold from 0.5 to 0.8
to reduce the number of neurons displayed and focus on few important neurons.
To see how differently the two subgroups are processed in the classifier,
Henry clicks on the neurons that are highly activated
only by the underperforming subgroup \#22 or
only by the well-performing subgroup \#296
to bring up the \conceptwindows (Figure~\ref{fig:neuronconcept}).
The \conceptwindows reveal that
the underperforming subgroup \#22 activates the neurons
that capture the \textit{text} and \textit{non-facial textures},
while the well-performing subgroup \#296 activates the neurons for \textit{mouths} and \textit{noses}.
This finding motivates Henry to wonder whether there may be some text or non-facial textures in the background of the images in the subgroup \#22
and decides to verify his conjecture by increasing the brightness of the images.
Indeed, as he expects, the brightened images 
have text and colored stripe patterns in the background (Figure~\ref{fig:subgroup22}),   associated with the lights, stands, and signs in the stadium.
Based on these findings,
Henry realizes that most of the images in the subgroup \#22
are for \textit{athletes in stadium}
and concludes that the classifier often misclassifies
\textit{smiling athletes in stadium}
as \textit{not smiling}.

\subsection{Model Performance Verification}

A common need in developing CNN image classifiers is 
to verify that they work as expected 
on both intended predictions 
and known undesirable cases~\cite{yeom2018privacy,li2019overfitting,farrand2020neither,zheng2020effects,KrishnakumarPSH21}. 
\toolname provides an interactive means for users to perform such verification.
For example,
a hypothetical CNN researcher Jane has prepared a biased CelebA dataset,
where she intentionally increases the co-occurrence
of the attribute \textit{black hair} and the label \textit{smiling}.
She expects that the model would potentially use image features related to \textit{black hair} to predict \textit{smiling}.

To verify her hypothesis, Jane launches \toolname.
As illustrated in Figure~\ref{fig:viscuit_ui},
Jane figures out that several image subgroups with low accuracies
are for the people with black hair at the first glance.
To see whether the model indeed uses the attribute \textit{black hair} for the predictions,
she clicks on subgroup \#14 with an accuracy of 36.4\%,
and \toolname displays subgroup \#380
that is similar to \#14
in terms of the last-layer features from the classifier
but has a high accuracy of 86.1\%.

In each of those subgroups,
clicking on an image brings up a \gradcamwindow.
It reveals that the classifier attends to \textit{forehead}, which is irrelevant to \textit{smiling},
for the underperforming subgroup \#14 (Figure~\ref{fig:viscuit_ui}-A1),
while for the images in the well-performing subgroup \#380,
the classifier attends to \textit{mouth},
which is closely related to \textit{smiling} (Figure~\ref{fig:viscuit_ui}-A2).
The confusion matrices
quantitatively summarize such misclassification
that many of the images of black-haired people are wrongly classified as \textit{smiling} even though they are not (Figure~\ref{fig:viscuit_ui}-A3).
Jane is now certain about her conjecture that 
the classifier often misclassifies
\textit{not smiling black-haired} people
as \textit{smiling}
due to the inappropriate attention to \textit{forehead}.

\section{Conclusion}
\label{sec:500conclusion}
We present \toolname,
a web-based interactive visualization tool
that helps users understand \textit{how} and \textit{why} CNN image classifiers are biased.
\toolname summarizes image subgroups with low accuracies
so that users can easily identify on what kinds of images the classifier underperforms
and select the subgroups to investigate more in depth. %
When users select an underperforming subgroup,
the \subgroupwindow of
\toolname displays
a well-performing subgroup that is similar to the selected underperforming subgroup in terms of the feature vectors from the classifier
and
confusion matrices for the two subgroups.
This can help users 
gain insights into 
the types of classification errors
and
the deviant features responsible for the biases.
Users can bring up \gradcamwindow{}s by clicking images
to learn which parts of an image have been deemed relevant for the classification.
Also,
from \neuronpathwindow,
users can figure out
the neurons and concepts 
responsible for misclassifications
and understand why the classifier performs unexpectedly poorly,
by clicking neurons and bringing up \conceptwindow{}s.
\toolname can be easily accessed through modern web browsers
and
is open-sourced
enabling easy extension of \toolname to various model architectures and datasets.
We believe \toolname would enhance people's understanding about CNN model biases
and accelerate practical applications of CNN image classifiers.

{\small
\bibliographystyle{ieee_fullname}
\bibliography{arxiv_main}
}

\end{document}